\documentclass[journal,twoside,web]{ieeecolor}
\usepackage{tmi}
\usepackage{cite}
\usepackage{amsmath,amssymb,amsfonts}
\usepackage{algorithmic}
\usepackage{graphicx}
\usepackage{textcomp}
\usepackage{algorithm}

\def\BibTeX{{\rm B\kern-.05em{\sc i\kern-.025em b}\kern-.08em
    T\kern-.1667em\lower.7ex\hbox{E}\kern-.125emX}}
\begin{document}
% Acronyms
% --------------------

% SSM model name
\newcommand{\ssmmodel}{Image2SSM}

% Shape representation
\newcommand\ShapeRepresentation{Radial-Basis-Function Shape}
\newcommand\shaperepresentation{radial basis function shape}
\newcommand\srepresentation{RBF-shape}
\newcommand\samploss{narrow band loss}

\newcommand\datasetone{femur}
\newcommand\datasettwo{liver}
\newcommand\datasetthree{right hip}

% Math
\newcommand\Real{\mathbb{R}}
\newcommand\diff{\mathbf{d}}
\newcommand\aP{\mathbf{\mathcal{P}}}
\newcommand\aN{\mathbf{\mathcal{N}}}
\newcommand\aD{\mathbf{\mathcal{D}}}
\newcommand\bN{\mathbf{N}}

%Colors
% --------------------
% \definecolor{cadmiumorange}{rgb}{0.93, 0.53, 0.18}
% \definecolor{celadon}{rgb}{0.67, 0.88, 0.69}
% \definecolor{darkolivegreen}{rgb}{0.33, 0.42, 0.18}
% \definecolor{darklavender}{rgb}{0.45, 0.31, 0.59}
\newcommand{\SE}[1]{{\color{blue}#1}}
\newcommand{\HX}[1]{{\color{orange}#1}}
\newcommand{\TODO}[1]{{\color{red}#1}}
\newcommand{\EDIT}[1]{{\color{purple}#1}}

% Example definitions.
% --------------------
\def\x{{\mathbf x}}
\def\L{{\cal L}}

\newcommand{\var}{{\rm var}}
\newcommand{\Tr}{^{\rm T}}
\newcommand{\vtrans}[2]{{#1}^{(#2)}}
\newcommand{\kron}{\otimes}
\newcommand{\schur}[2]{({#1} | {#2})}
\newcommand{\schurdet}[2]{\left| ({#1} | {#2}) \right|}
\newcommand{\had}{\circ}
\newcommand{\diag}{{\rm diag}}
\newcommand{\invdiag}{\diag^{-1}}
\newcommand{\rank}{{\rm rank}}
% careful: ``null'' is already a latex command
\newcommand{\nullsp}{{\rm null}}
\newcommand{\tr}{{\rm tr}}
\newcommand{\vech}{{\rm vech}}
\renewcommand{\det}[1]{\left| #1 \right|}
\newcommand{\pdet}[1]{\left| #1 \right|_{+}}
\newcommand{\pinv}[1]{#1^{+}}
\newcommand{\erf}{{\rm erf}}
\newcommand{\hypergeom}[2]{{}_{#1}F_{#2}}

% boldface characters
\renewcommand{\a}{{\bf a}}
\renewcommand{\b}{{\bf b}}
\renewcommand{\c}{{\bf c}}
\renewcommand{\d}{{\rm d}}  % for derivatives
\newcommand{\e}{{\bf e}}
\newcommand{\f}{{\bf f}}
\newcommand{\g}{{\bf g}}
\newcommand{\h}{{\bf h}}
%\newcommand{\k}{{\bf k}}
% in Latex2e this must be renewcommand
\renewcommand{\k}{{\bf k}}
\newcommand{\m}{{\bf m}}
\newcommand{\mb}{{\bf m}}
\newcommand{\n}{{\bf n}}
\renewcommand{\o}{{\bf o}}
\newcommand{\p}{{\bf p}}
\newcommand{\q}{{\bf q}}
\renewcommand{\r}{{\bf r}}
\newcommand{\s}{{\bf s}}
\renewcommand{\t}{{\bf t}}
\renewcommand{\u}{{\bf u}}
\renewcommand{\v}{{\bf v}}
\newcommand{\w}{{\bf w}}
\newcommand{\y}{{\bf y}}
\newcommand{\z}{{\bf z}}
%s\newcommand{\l}{\boldsymbol{l}}
\newcommand{\A}{{\bf A}}
\newcommand{\B}{{\bf B}}
\newcommand{\C}{{\bf C}}
\newcommand{\D}{{\bf D}}
\newcommand{\E}{{\bf E}}
\newcommand{\F}{{\bf F}}
\newcommand{\G}{{\bf G}}
\renewcommand{\H}{{\bf H}}
\newcommand{\I}{{\bf I}}
\newcommand{\J}{{\bf J}}
\newcommand{\K}{{\bf K}}
\renewcommand{\L}{{\bf L}}
\newcommand{\M}{{\bf M}}
\newcommand{\N}{\mathcal{N}}  % for normal density
\renewcommand{\O}{{\bf O}}
\renewcommand{\P}{{\bf P}}
\newcommand{\Q}{{\bf Q}}
\newcommand{\R}{{\bf R}}
\renewcommand{\S}{{\bf S}}
\newcommand{\T}{{\bf T}}
\newcommand{\U}{{\bf U}}
\newcommand{\V}{{\bf V}}
\newcommand{\W}{{\bf W}}
\newcommand{\X}{{\bf X}}
\newcommand{\Y}{{\bf Y}}
\newcommand{\Z}{{\bf Z}}

% this is for latex 2.09
% unfortunately, the result is slanted - use Latex2e instead
%\newcommand{\bfLambda}{\mbox{\boldmath$\Lambda$}}
% this is for Latex2e
\newcommand{\bfLambda}{\boldsymbol{\Lambda}}

% Yuan Qi's boldsymbol
\newcommand{\bsigma}{\boldsymbol{\sigma}}
\newcommand{\balpha}{\boldsymbol{\alpha}}
\newcommand{\bpsi}{\boldsymbol{\psi}}
\newcommand{\bphi}{\boldsymbol{\phi}}
\newcommand{\boldeta}{\boldsymbol{\eta}}
\newcommand{\Beta}{\boldsymbol{\eta}}
\newcommand{\btau}{\boldsymbol{\tau}}
\newcommand{\bvarphi}{\boldsymbol{\varphi}}
\newcommand{\bzeta}{\boldsymbol{\zeta}}

\newcommand{\blambda}{\boldsymbol{\lambda}}
\newcommand{\bLambda}{\mathbf{\Lambda}}
\newcommand{\bOmega}{\mathbf{\Omega}}
\newcommand{\bomega}{\mathbf{\omega}}
\newcommand{\bPi}{\mathbf{\Pi}}

\newcommand{\btheta}{\boldsymbol{\theta}}
\newcommand{\bpi}{\boldsymbol{\pi}}
\newcommand{\bxi}{\boldsymbol{\xi}}
\newcommand{\bSigma}{\boldsymbol{\Sigma}}

\newcommand{\bgamma}{\boldsymbol{\gamma}}
\newcommand{\bGamma}{\mathbf{\Gamma}}

\newcommand{\bmu}{\boldsymbol{\mu}}
\newcommand{\1}{{\bf 1}}
\newcommand{\0}{{\bf 0}}

\newcommand{\bs}{\backslash}
\newcommand{\ben}{\begin{enumerate}}
\newcommand{\een}{\end{enumerate}}

 \newcommand{\notS}{{\backslash S}}
 \newcommand{\nots}{{\backslash s}}
 \newcommand{\noti}{{\backslash i}}
 \newcommand{\notj}{{\backslash j}}
 \newcommand{\nott}{\backslash t}
 \newcommand{\notone}{{\backslash 1}}
 \newcommand{\nottp}{\backslash t+1}

\newcommand{\notk}{{^{\backslash k}}}
\newcommand{\notij}{{^{\backslash i,j}}}
\newcommand{\notg}{{^{\backslash g}}}
\newcommand{\wnoti}{{_{\w}^{\backslash i}}}
\newcommand{\wnotg}{{_{\w}^{\backslash g}}}
\newcommand{\vnotij}{{_{\v}^{\backslash i,j}}}
\newcommand{\vnotg}{{_{\v}^{\backslash g}}}
\newcommand{\half}{\frac{1}{2}}
\newcommand{\msgb}{m_{t \leftarrow t+1}}
\newcommand{\msgf}{m_{t \rightarrow t+1}}
\newcommand{\msgfp}{m_{t-1 \rightarrow t}}

\newcommand{\proj}[1]{{\rm proj}\negmedspace\left[#1\right]}
\newcommand{\argmin}{\operatornamewithlimits{argmin}}
\newcommand{\argmax}{\operatornamewithlimits{argmax}}

\newcommand{\dif}{\mathrm{d}}
\newcommand{\abs}[1]{\lvert#1\rvert}
\newcommand{\norm}[1]{\lVert#1\rVert}

%miscellaneous symbols
\newcommand{\ie}{{{i.e.,}}\xspace}
\newcommand{\eg}{{{\em e.g.,}}\xspace}
\newcommand{\EE}{\mathbb{E}}
\newcommand{\VV}{\mathbb{V}}
\newcommand{\sbr}[1]{\left[#1\right]}
\newcommand{\rbr}[1]{\left(#1\right)}
\newcommand{\cmt}[1]{}

%\title{Exploring Adaptivity and Correspondence in Particle-Based Statistical Shape Modeling}
\title{Adaptive Particle-Based Shape Modeling for Anatomical Surface Correspondence}
\author{Hong Xu and Shireen Y. Elhabian, \IEEEmembership{Member, IEEE}
\thanks{%This paragraph of the first footnote will contain the date on which
%you submitted your paper for review. 
Preprint. The National Institutes of Health supported this work under grant numbers NIBIB-U24EB029011 and NIAMS-R01AR076120. The content is solely the responsibility of the authors and does not necessarily represent the official views of the National Institutes of Health. }
\thanks{The corresponding author is Shireen Y. Elhabian. Both authors are with the Scientific Computing and Imaging Institute, Kahlert School of Computing, University of Utah, Salt Lake City, UT, USA \{hxu,shireen\}@sci.utah.edu }}
% \thanks{S. B. Author, Jr., was with Rice University, Houston, TX 77005 USA.
% He is now with the Department of Physics, Colorado State University,
% Fort Collins, CO 80523 USA (e-mail: author@lamar.colostate.edu).}
% \thanks{T. C. Author is with the Electrical Engineering Department,
% University of Colorado, Boulder, CO 80309 USA, on leave from the National
% Research Institute for Metals, Tsukuba, Japan (e-mail: author@nrim.go.jp).}}

\maketitle

\begin{abstract}
% These instructions provide guidelines for preparing papers for IEEE Transactions,
% but this version is specifically written to describe submission to IEEE TMI.
% Use this document as a template if you are using \LaTeX.
% Otherwise, use this document as an instruction set.
% The electronic file of your paper will be formatted further at IEEE.
% Paper titles should be written in uppercase and lowercase letters, not all uppercase.
% Avoid writing long formulas with subscripts in the title;
% short formulas that identify the elements are fine (e.g., "Nd--Fe--B").
% Keep the title short and do not write ``(Invited)'' in the title.
% Full names of authors are preferred in the author field, but are not required.
% Put a space between authors' initials. Only authors may appear in the author line
% of a manuscript. Authors are defined as individuals who have made an identifiable
% intellectual contribution to a manuscript to the extent that the individual can defend its contents.
% Define all symbols used in the abstract. Do not cite references in the abstract.
% Keep the abstract to 250 words or less.

Particle-based shape modeling (PSM) is a family of approaches that automatically quantifies shape variability across anatomical cohorts by positioning particles (pseudo landmarks) on shape surfaces in a consistent configuration. Recent advances incorporate implicit radial basis function representations as self-supervised signals to better capture the complex geometric properties of anatomical structures. 
However, these methods still lack self-adaptivity—that is, the ability to automatically adjust particle configurations to local geometric features of each surface, which is essential for accurately representing complex anatomical variability. 
This paper introduces two mechanisms to increase surface adaptivity while maintaining consistent particle configurations: (1) a novel neighborhood correspondence loss to enable high adaptivity and (2) a geodesic correspondence algorithm that regularizes optimization to enforce geodesic neighborhood consistency. 
We evaluate the efficacy and scalability of our approach on challenging datasets, providing a detailed analysis of the adaptivity-correspondence trade-off and benchmarking against existing methods on surface representation accuracy and correspondence metrics.
\end{abstract}


% File: aintroduction.tex
\section{Introduction}

Statistical Shape Modeling (SSM) is a collection of methods used to quantify shape variability across a population. In medical applications, SSM enables the analysis of morphological variation in anatomical and biological structures, enhancing our understanding of these forms and their associations with diseases and disorders \cite{atkins2017quantitative, bhalodia2020quantifying, LenzTalocruralJoint, VANBUURENHipOsteoarthritis, merle2019high,bruse2016statistical}. 
Two widely used shape representations in SSM condense shape variation into compact numeric forms for subsequent analysis: \textit{deformation fields} and \textit{landmarks}. Deformation fields define variation \textit{implicitly} through a lattice of transformations that deform regions between each shape in a population and a pre-defined or learned atlas. In contrast, landmarks use \textit{explicit} surface points that are evenly distributed and maintain a consistent configuration across shapes.
% Two popular shape representations are used to perform SSM, both of which work by condensing shape variation into a compact numeric representation and subsequently performing analysis on these representations: \textit{deformation fields} and \textit{landmarks}. The former defines variation \textit{implicitly} by using a lattice of transformations that deform regions between each shape in a population and a pre-defined (or learned) atlas, whereas the latter uses \textit{explicit} evenly spread surface landmarks that maintain the same configuration across shapes \cite{sarkalkan2014statistical}. 
Landmark-based representations, also known as point distribution models (PDMs), are preferred for analyzing surface variability due to their simplicity, computational efficiency, and interpretability \cite{sarkalkan2014statistical}. In contrast, deformation-based approaches are suited for modeling comprehensive image-wide correspondences but are generally less scalable.
% Landmark-based representations, also known as point distribution models (PDMs), are preferred for analyzing variability on shape surfaces due to their simplicity, computational efficiency, and interpretability. Deformation-based approaches are used in modeling comprehensive image-wide feature matches but are less scalable.
% Landmark-based representations, also known as point distribution models (PDMs), are preferred for analyzing variability on shape surfaces due to their simplicity, computational efficiency, and interpretability, while deformation-based approaches find use in modeling comprehensive image-wide feature matches but work at a more measured pace.
%

In this work, we focus on advancing landmark-based shape representations. Dense placement of corresponding landmarks is prohibitive to perform manually; advancements in computational methods, such as the minimum description length (MDL) approach \cite{davies2002minimum} and particle-based shape modeling (PSM) techniques \cite{cates2017shapeworks, oguz2016entropy}, have alleviated this burden by enabling automated landmark placement. This process generates PDMs that provide data-driven, objective, and reproducible characterizations of population-level variability. Although more scalable than deformation field methods, current PDM approaches can still require thousands of densely placed landmarks to capture subtle shape details, which slows down the optimization process, particularly in large-scale shape analysis studies.
%We focus on advancing the automatic placement of landmarks, which were initially manually annotated and subsequently automatically placed through computational methods such as minimum description length -- MDL \cite{davies2002minimum} and the family of particle-based shape modeling (PSM) approaches \cite{cates2017shapeworks, oguz2016entropy}. This automatic placement of particles creates PDMs, which, unlike manually annotated landmarks, are data-driven characterizations of population-level variabilities obtained in an objective, reproducible, and scalable manner. However, although more scalable than deformation field approaches, current PDM approaches still, at times, require thousands of densely placed landmarks to capture subtle shape detail faithfully, slowing down the optimization process.
%
Recent approaches have sought to reduce particle count requirements by distributing particles adaptively across the surface; specifically, particles are drawn to intricate or high-feature regions, thereby improving detail capture.
% Recent approaches have aimed to reduce particle count requirements by spreading them in a manner such that they adapt to the underlying surface, i.e., particles are attracted to intricate or high-feature regions, better capturing detail.
%
The first approach to address this was Image2SSM \cite{Image2SSM}, a deep learning-based method that introduced radial basis function shape representations (RBF-shapes) to make the PDM process shape-adaptive; this approach was later refined and extended through an optimization-based method in \cite{xu_nondeep_isbi}. These methods successfully adjusted particle positions to better conform to the surface but maintained an approximately uniform particle distribution, which constrained their representational capacity and scalability gains.
%
%The first approach to pursue this was Image2SSM \cite{Image2SSM}, a deep learning approach, which proposed using radial basis function shape representations (RBF-shapes) to make the PDM process shape adaptive and was later implemented and improved in an optimization method in \cite{xu_nondeep_isbi}. These methods successfully adjusted particle positions to better fit the surface but maintained roughly uniformly distributed particles, limiting their representative capacity and scalability gains.
%

This paper substantially extends our previous optimization method \cite{xu_nondeep_isbi} by 
(1) introducing a modified correspondence loss applied locally within particle neighborhoods to improve adaptivity, 
and (2) developing a geodesic correspondence regularization algorithm to correct correspondence irregularities. 
These advancements significantly enhance the adaptability of particles to local surface geometry while ensuring robust correspondence across the shape population. 
We rigorously evaluate our approach on challenging real datasets of \datasetone, \datasettwo, and \datasetthree\ anatomies. 
Our results demonstrate that our method achieves comparable or superior performance to \cite{xu_nondeep_isbi} as well as PSM \cite{cates2017shapeworks} in terms of mean and maximum two-way surface-to-surface distance and correspondence metrics even when PSM operates with twice the particle budget. 
The proposed methods will be incorporated into ShapeWorks \cite{cates2017shapeworks}, an open-source toolkit for automatically constructing PDMs from general anatomical structures.

\section{Methods}

\begin{figure}[!t]
\centerline{\includegraphics[width=\columnwidth]{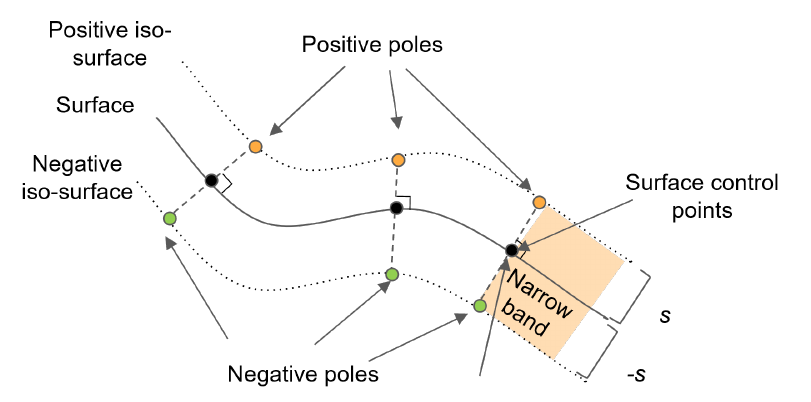}}
\caption{
Illustration of the global RBF shape model, where the surface is populated with control points and their respective dipoles, derived from surface normals. A system of equations is constructed from these control points and dipoles, enabling efficient querying of approximate distances for points near the surface.
%Shows how, for the global RBF shape, the surface is populated with control points and their respective dipoles are obtained from their normarls. We can build a system of equations using these control points and dipoles that allows querying of approximate distances to the surface of near surface points.
}\label{global_rbf}
\end{figure}

\begin{figure*}
    \centering
    \includegraphics[width=\textwidth, height=0.9\textheight, keepaspectratio]{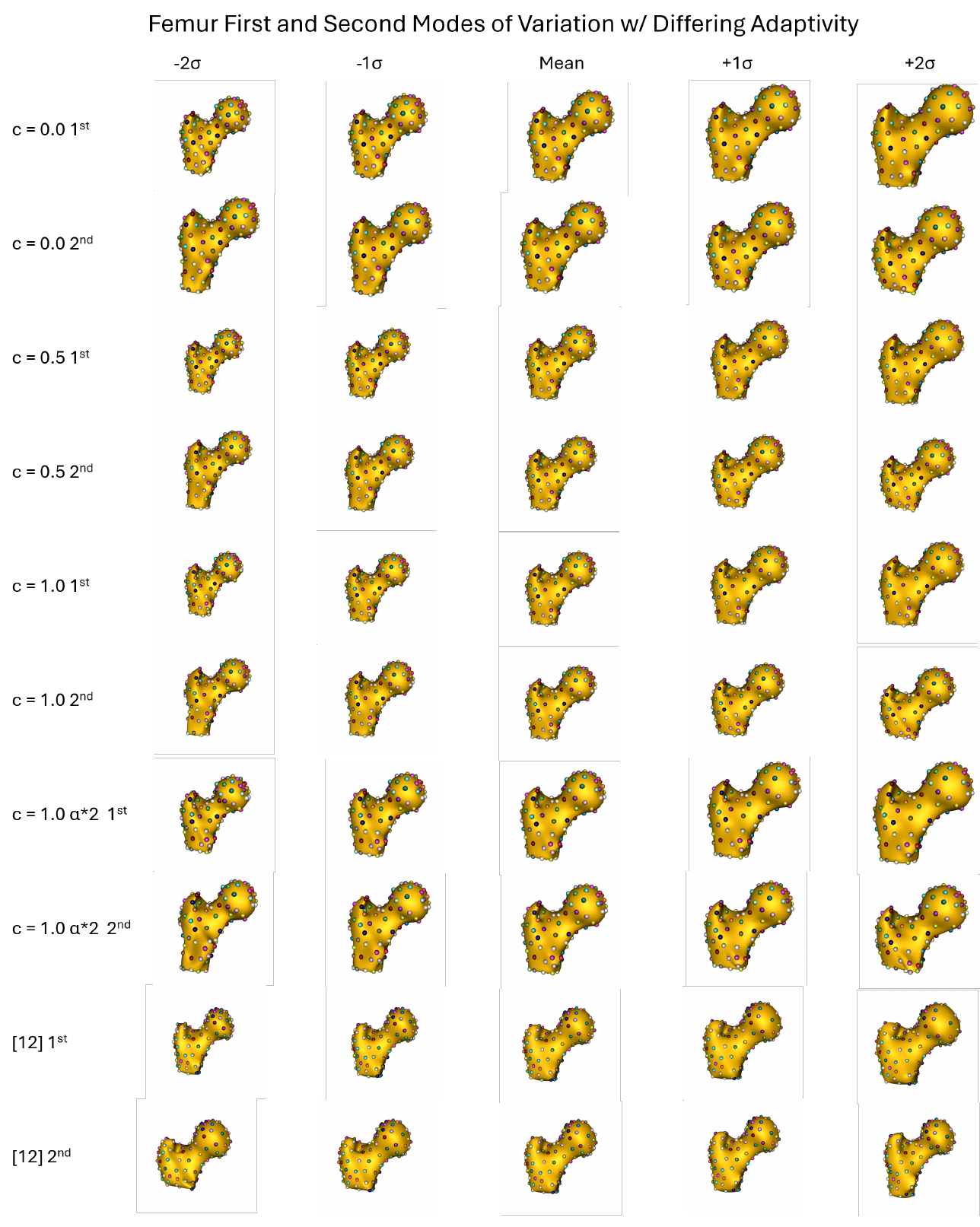}
    \caption{
        Each pair of columns shows the first and second modes of variation obtained from our method with different adaptivity weights $c$. For the next to last column, $\alpha = 10$, which is double the default of $\alpha = 5$. The last column shows the ones obtained from \cite{xu_nondeep_isbi}. From left to right, observe that adaptivity increases and more detailed features are captured, but correspondence deteriorates, which is evident in the surface artifacts. Compared to \cite{xu_nondeep_isbi}, our adaptivity degree is similar with $\alpha = 10$, but evaluation metrics are better as observed in Fig. \ref{fig:femur_metrics} and Fig. \ref{fig:femur_dist}.
    }
    \label{fig:adap_femurs}
\end{figure*}

An SSM encodes shape variation into a set of parameters designed to be as compact as possible. In the PSM formulation, this is achieved by placing particles to balance accurate geometric representation of individual surfaces (precise particle \textit{sampling}) with maintaining consistent landmark configuration across surfaces (particle \textit{correspondence} across shapes).

The proposed method optimizes a set of particles (or control points, as termed in RBF literature) to maintain sampling accuracy and correspondence, explicitly guided by the loss functions. These control points are placed on the surfaces of a cohort of shapes represented as binary segmentations, $\mathcal{S} = \{\S_i\}_{i=1}^I$, where $I$ denotes the number of shapes.
The optimization yields a collection of $J$ control points $\aP = \{ \P_i \}_{i=1}^I$ for each input shape, where the $i-$th shape PDM is denoted by
% If I add ^{(1)} to the following \P_i, some chapters turn blue, IDK why
$\P_i = [\p_{i,1}, \p_{i,2}, \cdots, \p_{i,J}]$ and $\p_{i,j} \in \mathbb{R}^3$.

The proposed approach is an optimization method that incorporates three distinct losses, each enforcing a specific characteristic for the final shape model: (1) particles are distributed on the surface to capture underlying geometric features accurately, (2) particles across shapes maintain consistent neighborhood configurations to ensure robust correspondences, and (3) a PCA on particle positions yields a compact representation of population-level variability, minimizing the number of required eigenvectors. These losses promote a balance between obtaining accurate geometric representations of surfaces in the population, good particle correspondence across shapes, and compact descriptions of the particle distribution, respectively.

Our approach also incorporates a novel geodesic correspondence regularization algorithm that promotes neighborhood particle correspondence across shapes and serves as a convergence criterion to achieve robust correspondence in the optimization. Additionally, this algorithm assists the optimization in escaping local minima arising from the non-convexity of the multi-loss objective function.

In this section, we present the background on RBF-shape, detail each layer of the optimization process alongside the proposed losses, and elaborate on the initialization and regularization strategies. Additionally, we provide implementation details and key considerations.

\begin{figure}
    \centering
    \includegraphics[width=\columnwidth, height=0.9\textheight, keepaspectratio]{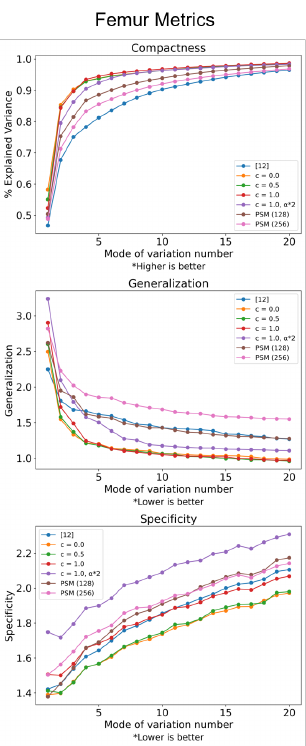}
    \caption{
        Compactness, generalization, and specificity of tested PDM approaches for femurs.
    }
    \label{fig:femur_metrics}
\end{figure}

\begin{figure}
    \centering
    \includegraphics[width=\columnwidth, height=0.9\textheight, keepaspectratio]{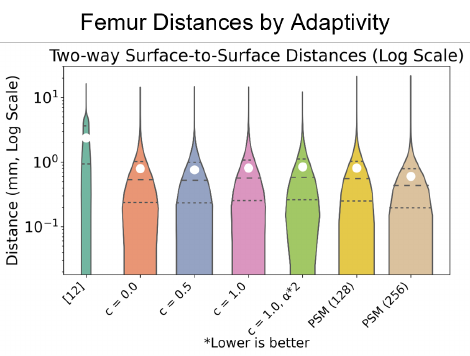}
    \caption{
        Two-way surface-to-surface distances on a log scale between original meshes and ones warped using particles for the femur dataset.
    }
    \label{fig:femur_dist}
\end{figure}

\begin{figure}
    \centering
    \includegraphics[width=\columnwidth, height=0.9\textheight, keepaspectratio]{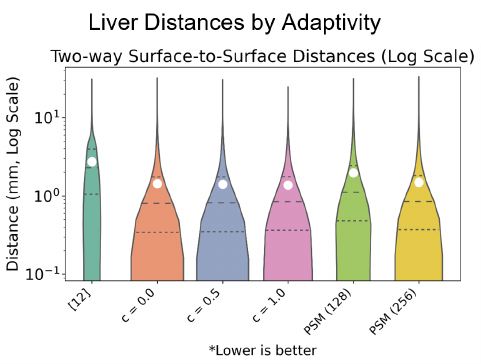}
    \caption{
        Two-way surface-to-surface distances on a log scale between original meshes and ones warped using particles for the liver dataset.
    }
    \label{fig:liver_dist}
\end{figure}

\begin{figure}
    \centering
    \includegraphics[width=\columnwidth, height=0.9\textheight, keepaspectratio]{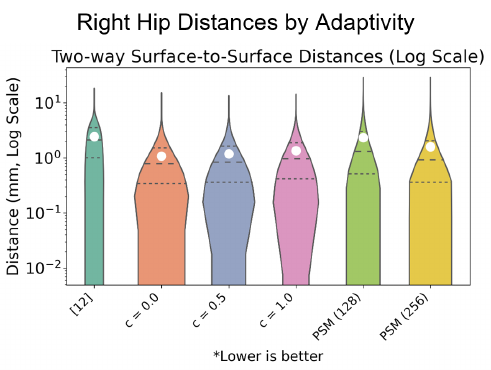}
    \caption{
        Two-way surface-to-surface distances on a log scale between original meshes and ones warped using particles for the hip dataset.
    }
    \label{fig:hip_dist}
\end{figure}

%%%%%%%%%%%%%%%%%%%%%%%%
% RBF Shape
%%%%%%%%%%%%%%%%%%%%%%%%

\subsection{Radial Basis Function Surface Representation}

RBF-shape is a method introduced by \cite{Turk1999VariationalIS} and \cite{rbf_surface1} that, given a set of control points (or particles) $\P_i$, constructs an implicit closed 3D surface of arbitrary topology. This method employs a linear system of radial basis functions to approximate the signed distance field around the surface, enabling efficient querying of near-surface distances. To define this signed distance field, off-surface points $\p_{i,j}^{(1)\pm} = \p_{i,j} \pm s\n_{i,j}$ (referred to as dipoles) are required for each control point. These dipoles are located at distances $s$ and $-s$ along the surface normal $\n_{i,j}$ from their respective control points. Fig. \ref{global_rbf} illustrates the configuration of control points and dipoles. The complete set of control points and dipoles is denoted $\widetilde{\P_i}$ for shape $i$, where $\widetilde{\P_i} = [\P_i, \P_i^{+}, \P_i^{-}].$

The linear system of radial basis functions can then be written as

% %\vspace{-0.18in}
\begin{align} \label{eq:rbf}
\begin{split}
    f_{\widetilde{\P_i}, \w_i}(\x) = \sum_{j \in \widetilde{\P_i}} w_{i,j} \phi(\x, \widetilde{\p}_{i,j}) + \mathbf{c}^T_i \x + c^0_{i} %+ \x^x c_{i,1} + \x^y c_{i,2} + \x^z c_{i,3},
\end{split}
% %\vspace{-0.2in}
\end{align}

\noindent where $\phi$ is any RBF basis function, such as the biharmonic $\phi(\x,\y) =|\x-\y|2$ or the triharmonic $\phi(\x,\y) =(|\x-\y|2)^3$, and $\mathbf{c}_i \in \mathbb{R}^3$ and $c^0_{i} \in \mathbb{R}$ encode the linear component of the surface. By solving the linear system to obtain $\w_i = [w_{i,1}, w_{i,2}, ..., w_{i,3J}, c_i^0, c_i^1, c_i^2,c_i^3] \in \mathbb{R}^{3J + 4}$ for $\x \in \widetilde{\P_i}$, 
% with regularization factors ensuring a fully determined system, 
we can query any $\x$ to compute a resolution-agnostic approximate signed distance to the surface. This compact and efficient signed distance querying method is used to assess reconstructed surface accuracy during particle optimization, as described in the definition of the sampling loss.

%%%%%%%%%%%%%%%%%%%%%%%%
% SSM Optimization
%%%%%%%%%%%%%%%%%%%%%%%%

\subsection{SSM Optimization}

Our optimization framework employs three loss functions, each of which will be detailed in the following.

% \vspace{0.1in}
% \noindent \textbf{Sampling loss:}

%%%%%%%%%%%%%%%%%%%%%%%%
% Sampling Loss
%%%%%%%%%%%%%%%%%%%%%%%%

\vspace{0.1in}
\subsubsection{Sampling Loss} \label{sampling}
The sampling loss is designed to configure particles so that the RBF shape reconstruction closely approximates the original shapes. Let $\B_i = [\b_{i,1}, ..., \b_{i,R}]$ represent $R$ randomly sampled near-surface points lying within a \textit{narrow band} at a distance of $\pm s$ from the surface. To achieve an even distribution, one approach is to minimize the distance between near-surface points and the closest particle. However, to enhance surface adaptivity, these points are evaluated using the RBF shape to obtain approximate distances to the reconstructed shape, which are then compared to the distances to the true surface, computed via a signed distance transform derived from the given binary segmentations. Any near-surface points where the distances between the reconstructed and true surfaces differ significantly indicate regions of poor representation, prompting an increase in attraction force on these points to encourage higher particle density in underrepresented areas.

To describe this distance minimization, let $\K^i \in \mathbb{R}^{R \times M}$ be the matrix containing the pairwise distances between each near-surface point $\b_{i,r}$ and each control point $\p_{i,j}$, such that, for the $i-$th shape, the $r,j-$th element is $k_{r,j}^i = \|\b_{i,r} - \p_{i,j}\|_2$. A soft minimum function is used so that each near-surface point influences all nearby control points, i.e., $\operatorname{softmin}(\K^i)$, where the $r,j$th element of $\operatorname{softmin}(\K^i)$ is $\exp{(-k_{r,j}^i)} / \sum_{j'=1}^J \exp (k_{r,j'}^i)$. This soft minimum is normalized over $\P_i$. Ultimately, we capture the RBF approximation squared error at the near surface points using $\e_i \in \mathbb{R}_+^{R}$, where $e_{i,r} = [f_{\widetilde{\P_i}, \w_i}(\b_{i,r}) - \D_i(\b_{i,r})]^2$, where $\D_i$ is the distance transform associated with the segmentation $\S_i$. Assuming $\E_i = \e_i \mathbf{1}_M^T$, where $\mathbf{1}_M$ is a ones-vector of size $M$, the sampling loss is written as: 

% %\vspace{-0.2in} 
\begin{align}
\begin{split}
  L^{sampl}_{\B_i, \D_i, \w_i} (\P_i, \bN_i) & = \\
  \operatorname{mean} & \left(\operatorname{softmin}(\K_i) \otimes \K_i \otimes  (c\E_i + (1-c) \J) \right)
\end{split}
% %\vspace{-0.2in} 
\end{align}

\noindent where $c$ is a weighting variable, $\otimes$ indicates the Hadamard (elementwise) multiplication of matrices, $\operatorname{mean}$ is the average over all the matrix elements, and $\J$ is a matrix of all ones. $c$ determines the inherent adaptivity desired in the application, with $c=1.0$ indicating full adaptivity and $c=0.0$ indicating uniform control point distribution across the surface.

%%%%%%%%%%%%%%%%%%%%%%%%
% Correspondence Loss
%%%%%%%%%%%%%%%%%%%%%%%%

\vspace{0.1in}
\subsubsection{Neighborhood Correspondence Loss} 
The neighborhood correspondence loss aims to match each particle $\p_{i,j}$'s neighborhood configuration on shape $i$ with the corresponding neighborhood of a template particle $\p_{t,j}$ on shape $t$, which is automatically selected based on cohort similarity using iterative closest point. This matching disregards translation, rotation, and scaling. Hence, the loss for each particle $j$ can be written as:

% %\vspace{-0.12in}
\begin{align}
\begin{split}
  L^{corres}_{j} (\P_1, ..., \P_K) & =\\ 
  & \sum_{k=1}^K \sum_{n=1}^{\mathcal{N}_6(\p_{t,j})} d \bigg( \T_{\p_{i,j}, \n_{i,j}, \mathcal{N}_6(\p_{t,j})}( \p_{i,n} ), \\
  & \T_{\p_{t,j}, \n_{t,j}, \mathcal{N}_6(\p_{t,j})}( \p_{t,n} ) \bigg)
\end{split}
% %\vspace{-0.25in}
\end{align}

\noindent where $d(\cdot, \cdot)$ represents the Euclidean distance, $N_q(\p_{i,j})$ denotes the indices of the $q$ nearest neighbors of particle $\p_{i,j}$ on shape $i$, and $\T_{\p_{i,j}, \n_{i,j}, \mathcal{N}_q(\p_{i,j})}$ is the transformation matrix that translates by $-\p_{i,j}$, rotates by $(1,0,0) - \n_{i,j}$, and scales based on the mean Euclidean distance between $\p_{i,j}$ and $\mathcal{N}_q(\p_{i,j})$. For implementation, this loss can be efficiently expressed in batch form and is scalable on GPU hardware.

The Frobenius norm-based correspondence loss previously used in \cite{xu_nondeep_isbi} enforced particle alignment across shapes by driving particles to similar positions, a method that is inadequate when handling shapes with substantial variations of differing scales. The effects of this loss are analyzed in the results section.

%%%%%%%%%%%%%%%%%%%%%%%%
% Eigenshape loss
%%%%%%%%%%%%%%%%%%%%%%%%

\vspace{0.1in}
\subsubsection{Eigenshape Loss}

The eigenshape loss encourages particles to align with the principal modes of variability, as first described in \cite{EigenshapeKOTCHEFF1998303}. This loss is mathematically equivalent to minimizing the Mahalanobis distance within the particle system, ensuring that shape variation is concentrated within the dominant PCA modes.

Given a minibatch of size $K$, the eigenshape loss is the differential entropy $H$ of the samples in the minibatch. Assuming a Gaussian distribution, it can be written as:

% %\vspace{-0.2in}
\begin{align}
\begin{split}
  L^{eigen}_{\boldsymbol{\mu}} (\P_1, ..., \P_K) & = H(\P)  \\% = \frac{1}{2} \log |\Sigma| \\ 
   = \frac{1}{2} \log \left| \frac{1}{3JK}\sum_{k=1}^K \right.& \left. \left(\P_k - \boldsymbol{\mu}\right) \left(\P_k - \boldsymbol{\mu}\right)^T \right|
\end{split}
% %\vspace{-0.2in}
\end{align}

\noindent where $\P$ indicates the random variable of the particles in shape space and $|\cdot|$ is the matrix determinant.

The neighborhood correspondence and eigenshape losses are computed starting from the second epoch, following a lagging strategy in which the first epoch serves as a burn-in stage to establish an initial set of correspondences across the shapes.

%%%%%%%%%%%%%%%%%%%%%%%%
% Total Loss
%%%%%%%%%%%%%%%%%%%%%%%%

\vspace{0.1in}
\subsubsection{Total Loss} 

The total loss is optimized over minibatches of size $K$ and is defined as the sum of the three individual losses.

% %\vspace{-0.3in} 
\begin{align}
\begin{split}
  L_{\mathcal{I}, \aD, \partial \aD}(\aP_K, \aN_K) &= \sum_{i=1}^K \biggl( \alpha L^{sampl}_{\B_i, \D_i, \w_i} (\P_i, \bN_i) \biggl) \\
  & +  \beta L^{eigen}_{\boldsymbol{\mu}} (\P_1, ..., \P_K)\\
  & + \gamma L^{corres}_{\boldsymbol{\mu}} (\P_1, ..., \P_K)
\end{split}
% %\vspace{-0.2in} 
\end{align}

\noindent where $\alpha, \beta$, and $\gamma \in \mathbb{R}_+$ are the hyperparameters for each loss, $\aP_K$ and $\aN_K$ represent the particles and their surface normals in the minibatch, respectively, $\aD = \{\D_i\}{i=1}^I$ denotes the distance transform for each shape, and $\partial \aD = {\partial \D_i}_{i=1}^I$ are the partial derivatives of each signed distance field, from which the normals at any point can be queried.

%%%%%%%%%%%%%%%%%%%%%%%%
% Optimization details
%%%%%%%%%%%%%%%%%%%%%%%%

\subsection{Optimization Details}

This section addresses essential aspects of the optimization process required for effective regularization and stability, along with a detailed examination of the initialization strategy.

%%%%%%%%%%%%%%%%%%%%%%%%
% Geodesic correspondence algorithm
%%%%%%%%%%%%%%%%%%%%%%%%

\vspace{0.1in}
\subsubsection{Geodesic Correspondence Algorithm} This algorithm adjusts the correspondence of particle systems inspired by the process of manual inspection of correspondence. It mimics the process of looking at a reference shape and ensuring that every other shape maintains a similar geodesic neighborhood structure for each particle.

Shown in Algorithm \ref{alg:geodesic_corr}, the algorithm starts by computing the geodesic neighborhood for each particle of the reference shape using the shape's mesh structure to compute geodesic distance. A particle $j$ is considered a neighbor if its geodesic distance is within a factor of $1.5$ of particle $j$'s closest geodesic neighbor distance. Then, for each non-reference shape, we identify for each particle $j$, how many neighbors do not match the neighborhood of particle $j$ in the reference shape. In order of most mismatched neighbors to least, we update particle $j$ to be at the same distance as the neighbor $j_{neighbor}$'s closest geodesic neighbor on the geodesic path from particle $j$ to $j_{neighbor}$.

The asymptotic runtime of this algorithm is $O(I \cdot J \cdot V \log V)$, where $\V$ is the maximum number of vertices for all shapes and considering fully triangular meshes. This algorithm can fix minor to moderate correspondence issues. It may also serve as an algorithmic verification method for good correspondence given that the number of particles with neighborhood discrepancies on a shape and the particle configuration similarity between it and the reference shape are indicative of correspondence quality. As such, we are able to use it as a convergence criterion for our experiments. This algorithm is applied in intervals of 25 epochs to allow the optimization to reattain even spreads of particles on each shape, which is required for the algorithm to be effective. 

\begin{algorithm}[H]
\caption{Geodesic Correspondence Algorithm}\label{alg:geodesic_corr}
\begin{algorithmic}[1]
\REQUIRE $particle\_system$: array of particle sets for each shape
\REQUIRE $mesh\_structures$: mesh data
\REQUIRE $reference\_id$: index of reference shape

\STATE $reference\_particles \leftarrow particle\_system_{reference\_id}$
\STATE Compute $reference\_neighborhoods$, geodesic neighbors of each particle in $reference\_particles$

\FOR{each shape $i$ in the dataset $\neq reference\_id$}

    \STATE (a) Identify particle indices in shape $i$ according to how many neighbors do not maintain the same configuration.

    \STATE (b) Order these indices from most neighborhood discrepancy to least.

    \STATE (c) In (b) order, move particles by geodesically walking towards neighbors. Update $particle\_system$.

\ENDFOR

\RETURN $particle\_system$
\end{algorithmic}
\end{algorithm}

%%%%%%%%%%%%%%%%%%%%%%%%
% Snapping to Surface
%%%%%%%%%%%%%%%%%%%%%%%%

\vspace{0.1in}
\subsubsection{Surface Snapping} The sampling objective can push particles off the surface if the mean of the narrow band points closest to a particle is off the surface. Additionally, the correspondence and eigenshape functions have no constraints for particles to remain on the surface. Thus, to constrain the particles to the shape surfaces, we use the distances from $\aD$ multiplied by the unit normals from $\partial \aD$ to snap particles to the surface after each iteration. This guarantees that the particles will always be on the surface, which is cheaper than an iterative Newton-Raphson method used in \cite{cates2017shapeworks} and more accurate than relying on the sampling loss alone as in \cite{xu_nondeep_isbi}.

%%%%%%%%%%%%%%%%%%%%%%%%
% Initialization step
%%%%%%%%%%%%%%%%%%%%%%%%

\vspace{0.1in}
\subsubsection{Initialization Step} 
We introduce a particle initialization technique that uses farthest point geodesic sampling to obtain evenly spaced particles across mesh surfaces and the Hungarian algorithm with geodesic distances to initialize with mild matching correspondence against a reference shape. 

Farthest point geodesic sampling \cite{geodesic_farthest} is a technique used to generate a set of well-spaced sample points on a 3D surface, ensuring maximal coverage by iteratively selecting points that are maximally distant from previously chosen ones in terms of geodesic (surface) distance. In contrast to random or grid-based sampling, which can lead to clusters or irregular spacing, farthest point sampling generates points that are as evenly spread out as possible across the surface. Obtaining evenly spaced particles reduces the epoch count necessary to achieve even spread during optimization.

The Hungarian algorithm \cite{Kuhn1955Hungarian} is a combinatorial optimization algorithm designed to solve the assignment problem, which seeks the optimal way to assign each element in one set to exactly one element in another set based on a cost function, thereby achieving one-to-one matching. This characteristic makes it ideal for applications requiring unique, one-to-one correspondences, such as job assignments, point correspondences in shape analysis, or feature matching in computer vision. Given a pairwise distance metric between matches, it is proven to provide an optimal solution in $O(n^3)$ time, where $n$ is the number of elements. By procuring a solution by giving pairwise geodesic distances, we procure a reasonable initial matching of particles obtained by farthest point geodesic sampling.

Lastly, given a particle system obtained from the above procedure, we run Algorithm \ref{alg:geodesic_corr} on it to refine the correspondence.

% One observed issue was that sometimes particles maintain good Euclidean neighborhood correspondence but not geodesic, which is desired when operating on shape surfaces. However, once the particles attain geodesic correspondence, the neighborhood correspondence loss can keep geodesic correspondence in most cases.

% Starting from randomly sampled particles from a template's surface projected to all others, for each level, we start the optimization with 100 epochs of \textit{initialization}, which consists of a uniformly-sampled particle optimization to get particles evenly spread, and a geodesic correspondence routine runs to enforce matching geodesic neighborhood configurations.

% This is necessary because particles might maintain good Euclidean neighborhood correspondence but not geodesic, which is desired when operating on shape surfaces. However, once the particles attain geodesic correspondence, the neighborhood correspondence loss can keep geodesic correspondence in most cases.
% \section{Experiments}

\begin{figure*}
    \centering
    \includegraphics[width=\textwidth, height=0.9\textheight, keepaspectratio]{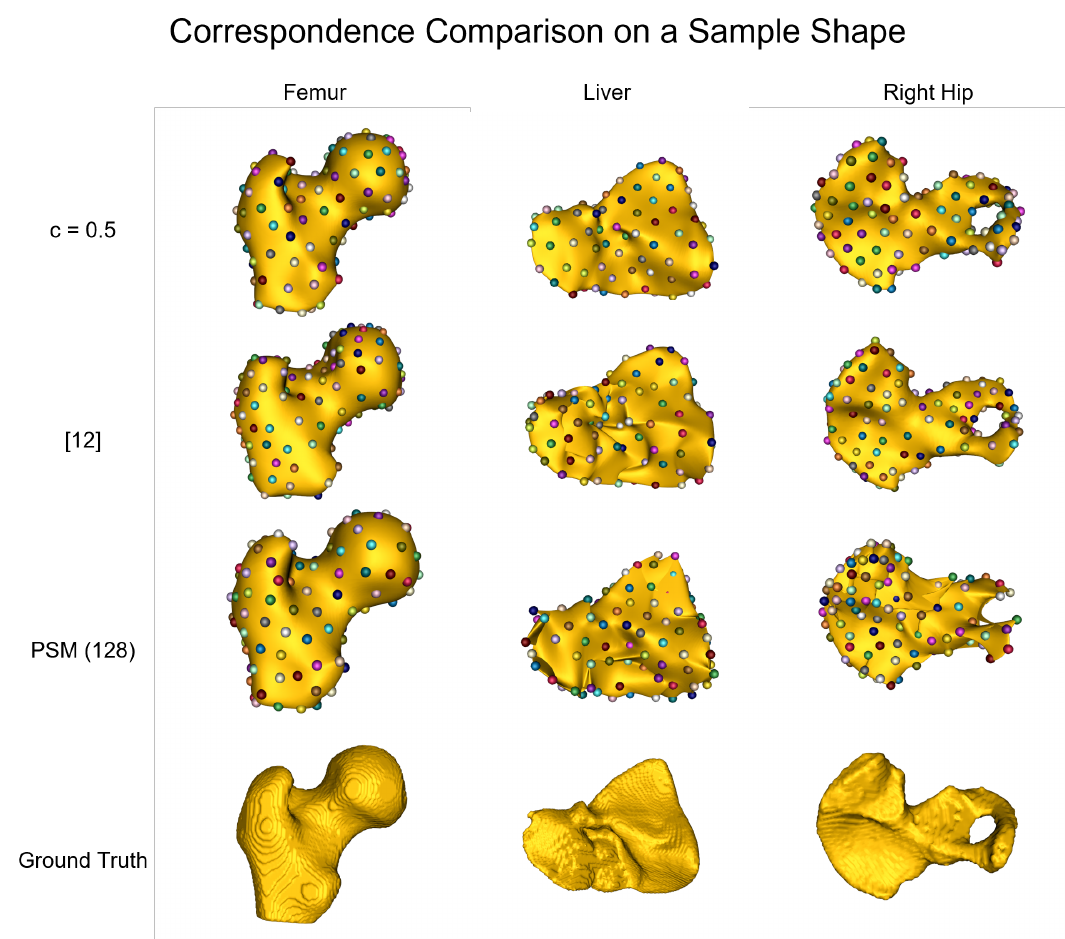}
    \caption{
        Shows sample shapes from each dataset with their individual warped reconstructions and the ground truth shape. Good correspondence manifests as reconstructed surfaces without spurious artifacts and folds absent in the ground truth shape. Excellent correspondence results and measured adaptivity are observed with the proposed approach where $c = 0.5$.
    }
    \label{fig:corr_samples}
\end{figure*}

\begin{figure}
    \centering
    \includegraphics[width=\columnwidth, height=0.9\textheight, keepaspectratio]{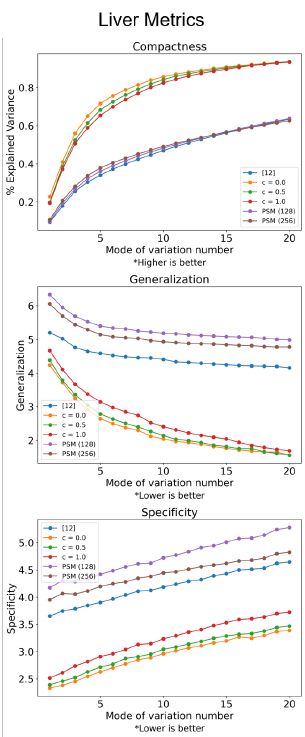}
    \caption{
        Compactness, generalization, and specificity of tested PDM approaches for livers.
    }
    \label{fig:liver_metrics}
\end{figure}

\begin{figure}
    \centering
    \includegraphics[width=\columnwidth, height=0.9\textheight, keepaspectratio]{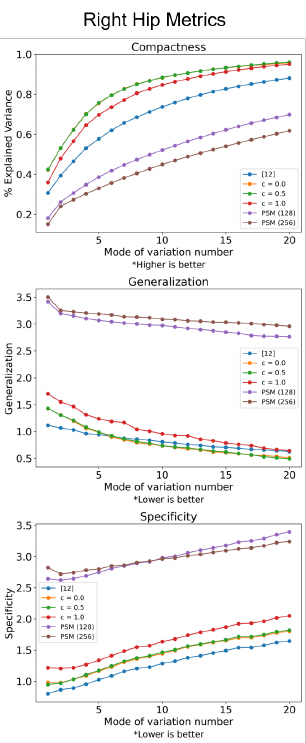}
    \caption{
        Compactness, generalization, and specificity of tested PDM approaches for hips.
    }
    \label{fig:hip_metrics}
\end{figure}

\begin{figure}
    \centering
    \includegraphics[width=\columnwidth, height=0.9\textheight, keepaspectratio]{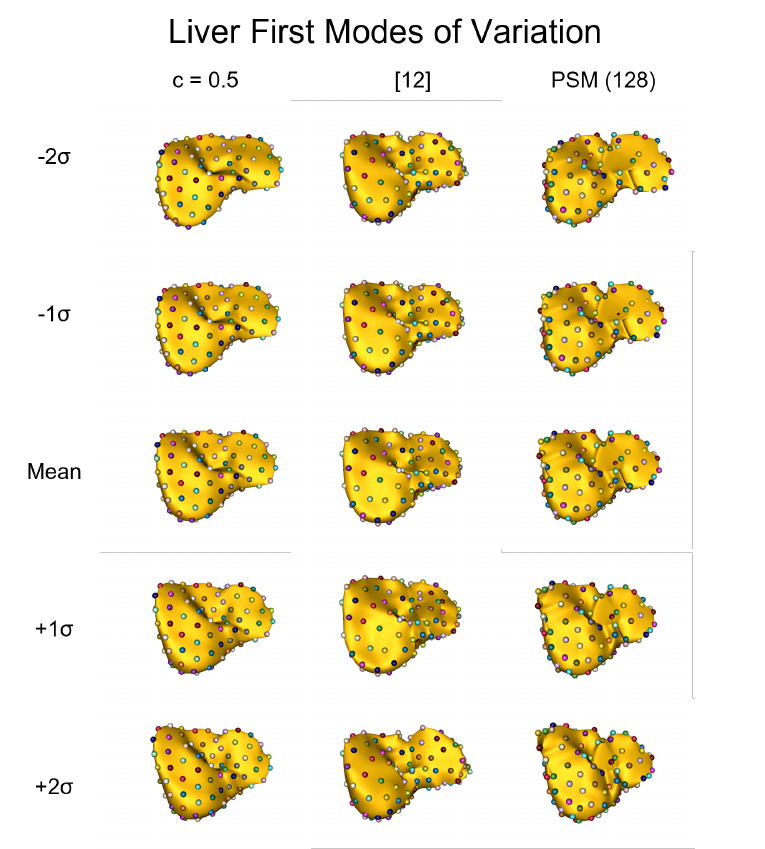}
    \caption{
        The first mode of variation for the liver dataset for different strategies. The geodesic correspondence algorithm handles missed correspondences at the thin edges of the shape, which also aids in spreading particles more evenly.
    }
    \label{fig:liver_modes}
\end{figure}

\begin{figure}
    \centering
    \includegraphics[width=\columnwidth, height=0.9\textheight, keepaspectratio]{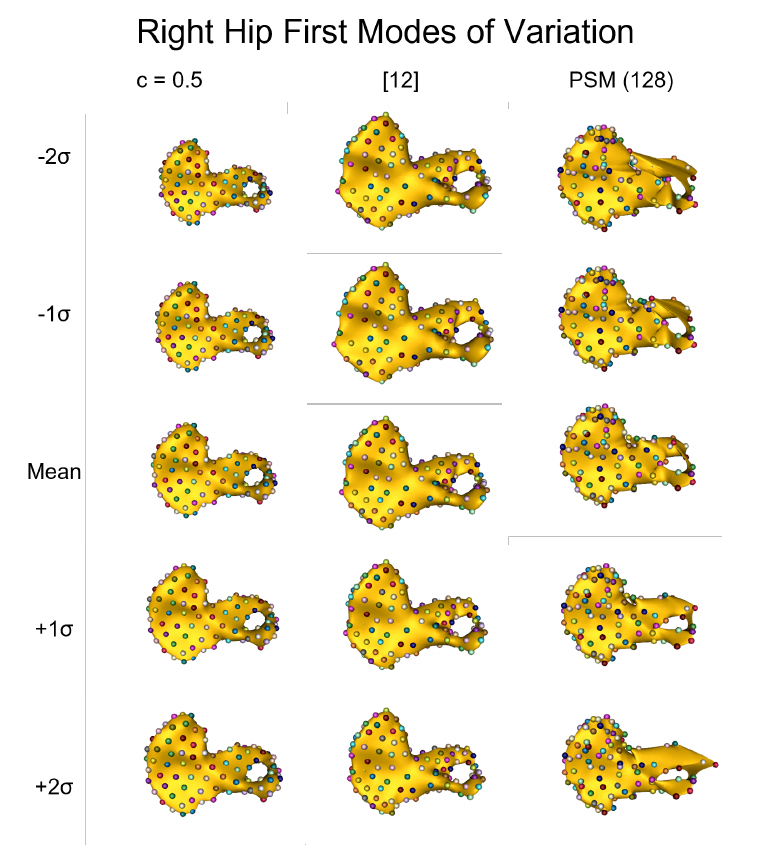}
    \caption{
        The first mode of variation for the right hip for different strategies. The geodesic correspondence algorithm for the proposed approach procures better results without artifacts than \cite{xu_nondeep_isbi} and PSM.
    }
    \label{fig:hip_modes}
\end{figure}
% File: aresults.tex
\section{Results}

In this section, we evaluate the impact of adaptivity on surface-to-surface distance and shape correspondences. Additionally, we test and compare the effectiveness of our approach on three real datasets against the open source version of PSM \cite{cates2017shapeworks} and \cite{xu_nondeep_isbi} by evaluating the relevant metrics to surface representation accuracy and model quality.

%%%%%%%%%%%%%%%%%%%%%%%%
% Datasets
%%%%%%%%%%%%%%%%%%%%%%%%

\subsection{Evaluation Metrics}

 We evaluate our approach based on the following metrics: 
 
 \subsubsection{Two-way surface-to-surface distance} This method evaluates distances between the ground-truth meshes and reconstructed meshes, obtained by warping a reference original mesh to specific samples using interpolated transformations obtained from corresponding particle positions. The vertex-to-surface distances between ground-truth and reconstructed meshes and vice versa are concatenated to obtain the distances for each vertex on the ground-truth and reconstructed meshes, thus two-way surface-to-surface distance. 
 
 These distances are indicative of surface representation and correspondence quality. Lower distances indicate that corresponding particles capture variation across the whole surface correctly, even between particles via interpolation. Conversely, higher distances indicate failure to capture important variability due to insufficiently dense sampling or a missed correspondence (missed correspondences result in artifacts when warped). 
 
 \subsubsection{Compactness} Compactness measures the percentage of variance captured per mode of variation, which is an eigenvector in our case of PCA analysis. Higher compactness over fewer modes of variation is desirable because variation is explained more concisely. 

 \subsubsection{Generalization} Refers to the ability to represent unseen shape instances. It is computed using a leave-one-out strategy and evaluating the distance (over principal eigenvectors) of the excluded sample to the others. Lower generalization indicates the capability to represent unseen shapes.
 
 \subsubsection{Specificity} The ability to generate realistic shapes when sampling from the model. It is evaluated by sampling 25000 particle set samples uniformly from the PCA model and computing their distance to the closest particle set in the dataset. Lower specificity signifies a lower likelihood of including spurious or unrealistic samples in our model.

 Combined, these four metrics are indicative of the overall quality of PDMs.

%%%%%%%%%%%%%%%%%%%%%%%%
% Datasets
%%%%%%%%%%%%%%%%%%%%%%%%

\subsection{Datasets}

The first dataset consists of binary segmentations from 40 proximal femur CT scans devoid of pathologies \cite{femurAnupama}. Left femurs are reflected so they are all aligned as right femurs. The second dataset consists of 100 liver meshes obtained from \cite{Ma-2021-AbdomenCT-1K} and processed using \cite{cates2017shapeworks}. The liver dataset is challenging due to the high variability in shapes. The third dataset is formed by 100 right hip binary segmentations obtained from the TotalSegmentator MR images dataset \cite{total_segmentator1, total_segmentator2} and processed with \cite{cates2017shapeworks}. The right hip dataset is challenging due to its small features and thin structure.

%%%%%%%%%%%%%%%%%%%%%%%%
% Implementation Details
%%%%%%%%%%%%%%%%%%%%%%%%

\subsection{Implementation Details}

The optimization is performed in two stages. The first stage finds a good correspondence uniformly distributed particles system using $c = 0.0$ and algorithm \ref{alg:geodesic_corr} performed on the particles every 25 epochs. The second stage activates adaptivity with $c \in (0,1]$ and without algorithm \ref{alg:geodesic_corr}. This two-stage approach ensures particle systems with acceptable correspondence results. Surface snapping was performed every epoch.

The Autograd functionality of PyTorch 1.12.1 was used to automatically backpropagate losses using the SGD optimizer. The biharmonic kernel was used as the basis function. 
The hyperparameters used for the first stage were $learning\_rate = 1$ and $\alpha = 10$; $\beta = 0.05$ and $\gamma = 5$ for liver, $\beta = 0.01$ for right hip and femur, and $\gamma = 0.1$ for right hip and $\gamma = 0.5$ for femur. Convergence was determined by setting a tolerance for the number of particles with mismatched geodesic neighborhoods for each shape in algorithm \ref{alg:geodesic_corr}. In the second stage, only $\alpha = 5$ and the desired adaptivity $c \in (0,1]$ were changed. The second stage ran for 200 epochs, at which point particle movement had reduced to a small tolerance for every dataset.

%%%%%%%%%%%%%%%%%%%%%%%%
% Adaptivity Results
%%%%%%%%%%%%%%%%%%%%%%%%

\subsection{Adaptivity Results}

Our first set of experiments aims to showcase the adaptivity capabilities enabled by the neighborhood correspondence loss and explore how the adaptivity weight $c$ interacts with distance and correspondence metrics results. These experiments are performed on femur since it facilitates the observation of the relevant phenomena qualitatively and quantitatively. 

Fig. \ref{fig:adap_femurs} shows the adaptivity results on the femur dataset through its first mode of variation. The particles better adapt to the surfaces as adaptivity weight $c$ increases, capturing finer detail of the shape. To increase adaptivity even further, we can increase $\alpha$ (doubled $\alpha$ shown). This adaptivity is enabled by using the neighborhood correspondence loss instead of the Frobenius loss since the former does not restrict particle positioning to be biased toward the mean.

Although adaptivity can aid in capturing surface detail and decrease surface-to-surface distance, too much adaptivity is detrimental to correspondence because particles might be attracted to noisy or highly variable features. In the case of the femur, as particles move away from flat areas, they can populate differing feature rich areas. This deteriorating correspondence can be seen in Fig. \ref{fig:femur_dist} and Fig. \ref{fig:femur_metrics}. Observe that $c = 0.5$ maintains similar correspondence metrics to $c = 0.0$ and sees improvement in mean surface-to-surface distance. However, as adaptivity grows further, both the correspondence metrics and surface-to-surface distances suffer due to poorer correspondence. 

% \begin{figure}
% \includegraphics[width=0.48\textwidth]{figures/adap_femur_s2s.png}
% \caption{
% Two-way surface-to-surface distances and logarithm of two-way surface-to-surface distances between original meshes and ones warped using particles.
% }\label{fig:adap_femur_s2s}
% \end{figure}

% \begin{figure}
% \includegraphics[width=0.48\textwidth]{figures/adap_femur_metrics.png}
% \caption{
% Compactness, generalization and specificity of optimizations with different adaptivity weights $c$. $\alpha$*2 indicates $\alpha = 10$.
% }\label{fig:adap_femur_metrics}
% \end{figure}

%%%%%%%%%%%%%%%%%%%%%%%%
% Comparison Shape Model Quality
%%%%%%%%%%%%%%%%%%%%%%%%

\subsection{Shape Model Quality}

The second set of experiments compares our optimization against \cite{xu_nondeep_isbi} and PSM \cite{cates2017shapeworks} in terms of mean and maximum two-way surface-to-surface distance and correspondence metrics on three real datasets. We also test against PSM with double the particle budget to evaluate whether our approach can obtain similar results with a reduced particle budget. A concerted effort was made to find the optimal PSM hyperparameters to attain a balance between sampling and correspondence for each dataset.

We provide model quality comparisons for each dataset.
% Femur
The femur distance graphs are given in Fig. \ref{fig:femur_dist} and metrics are given in Fig. \ref{fig:femur_metrics}. 
% Liver
Liver model distances are shown in Fig. \ref{fig:liver_dist} and metrics are given in Fig. \ref{fig:liver_metrics}. Fig. \ref{fig:liver_modes} shows qualitative visualizations of the modes of variation for the liver dataset. 
% Hip right
For the right hip dataset, distances are shown in Fig. \ref{fig:hip_dist} and metrics in Fig. \ref{fig:hip_metrics}. Fig. \ref{fig:hip_modes} shows qualitative visualizations of the modes of variation for the right hip dataset. 

The proposed method with $c = 0.5$ yields better or comparable maximum and mean surface-to-surface distance and correspondence metrics than all other tested methods, even against PSM with higher particle counts. An exception is observed on femurs with the mean surface-to-surface distance compared to PSM with 256 particles, where a thin margin outperforms it. However, the proposed method still outperforms PSM with  128 particles, indicating a particle budget advantage below double.

The improvements in surface-to-surface distance over \cite{xu_nondeep_isbi} can be directly attributed to the inclusion of surface snapping. The improvements in correspondence metrics are due to improved particle configuration matching achieved through the geodesic correspondence algorithm. Fig. \ref{fig:corr_samples} shows cohort samples for each evaluated dataset to showcase the correspondence improvements qualitatively. Given good particle configuration matches, the eigenshape loss refines the PDM by promoting compact, general, and specific eigenspace properties. Together, the techniques introduced provide a more ample toolset for optimizing particle-based shape models as they will be subsequently incorporated into PSM.

%An exception is observed on femurs with the mean surface-to-surface distance compared to PSM with 256 particles, where a thin margin outperforms it.

% Concerning correspondence metrics, the proposed method performs on par or better than other approaches, except for specificity against both PSM femur with 128 and 256 particles, where a thin margin outperforms it. The proposed approach and PSM achieve good correspondence on the femur dataset, making these small discrepancies due to the differing sampling styles where the proposed approach uses the surface to attract particles, whereas PSM repels particles from each other.
% File: aconclusion.tex
\section{Conclusion}

We demonstrate an approach for improving PDM construction that improves on previously proposed optimization ideas to increase adaptivity and correspondence quality. Our novel losses and the geodesic correspondence algorithm offer new insights regarding developing correspondence models that better capture variability while being faithful to the surface representations. Future work involves exploring efficient geodesic correspondence-based losses to improve correspondence results further, and hierarchical particle optimizations that include global and local optimizations to further improve particle budget usage.

\bibliographystyle{IEEEbib}
{\footnotesize \bibliography{references}}

\end{document}